\documentclass[conference]{IEEEtran}
\IEEEoverridecommandlockouts
\usepackage{cite}
\usepackage{url}
\usepackage{amsmath,amssymb,amsfonts}
\usepackage{algorithmic}
\usepackage{graphicx}
\usepackage{textcomp}
\usepackage{xcolor}
\usepackage{booktabs}   
\usepackage{array}      
\usepackage{multirow}   
\usepackage{caption}    
\usepackage{array}
\usepackage{makecell}   
\usepackage{colortbl}   
\usepackage{tcolorbox}
\usepackage{rotating}
\usepackage{adjustbox}
\usepackage{nicematrix}  
\usepackage{pifont}
\usepackage{enumitem}
\def\BibTeX{{\rm B\kern-.05em{\sc i\kern-.025em b}\kern-.08em
    T\kern-.1667em\lower.7ex\hbox{E}\kern-.125emX}}
\begin{document}

\title{Formalizing Heterogeneous Interaction Modeling for Multi-Agent Debate in Visual Geo-Localization}

\author{
    \IEEEauthorblockN{
        Zijie Zhou\textsuperscript{1,2}, 
        Dandan Zhu\textsuperscript{1,2,}\IEEEauthorrefmark{1}, 
        Heng Zhang\textsuperscript{3}, 
        Yanming Yang\textsuperscript{4}, 
        Xu Liu\textsuperscript{1,2}, 
        Huishen Jiao\textsuperscript{1,2}
    }
    \vspace{0.5cm} 
    
    \IEEEauthorblockA{\textsuperscript{1}China University of Petroleum (Beijing), Beijing, China}
    \IEEEauthorblockA{\textsuperscript{2}Hainan Institute of China University of Petroleum (Beijing), Sanya, Hainan, China}
    \IEEEauthorblockA{\textsuperscript{3}South China Normal University, Foshan, China \quad \textsuperscript{4}Wuhan University, Wuhan, China}
    
    \thanks{\IEEEauthorrefmark{1} Corresponding author: Dandan Zhu (e-mail: zhu.dd@cup.edu.cn).}
    \thanks{This research was supported by Oil \& Gas Major Project (No. 2025ZD1404600), CNPC Innovation Found (No. 2022DQ02-0609), and Frontier Interdisciplinary Exploration Research Program of China University of Petroleum, Beijing (Grant No. 2462024XKQY003).}
}

\maketitle


\begin{abstract}
    Visual geo-localization requires complex reasoning to infer image locations without GPS metadata. Traditional retrieval methods are limited by database quality while Large Vision Language Models often struggle with diverse scenes. Although existing multi-agent systems improve collaboration, they treat interactions uniformly and lack effective conflict resolution. We propose GraphGeo, a multi-agent debate framework based on heterogeneous graph neural networks for visual geo-localization. The framework models diverse debate relationships through typed edges and separates supportive collaboration, competitive argumentation and knowledge transfer. We introduce a dual-level debate mechanism combining node-level refinement and edge-level argumentation modeling. A cross-level topology refinement strategy enables co-evolution between graph structure and agent representations. Experiments on multiple benchmarks demonstrate GraphGeo outperforms single model and multi-agent baselines. The framework converts cognitive conflicts between agents into higher geo-localization accuracy through structured debate.
\end{abstract}

\begin{IEEEkeywords}
Large Visual Geo localization, Multi agent, Debate, Heterogeneous Graph Neural Networks, Large Vision Language Models
\end{IEEEkeywords}

\section{Introduction}
Visual geo localization aims to identify the geographic position of an image without GPS metadata. This task plays an important role in applications that include analysis of historical human mobility, support for autonomous navigation and location aware recommendation services~\cite{hays2008im2gps,baatz2012large}. Traditional methods address this problem by retrieving similar images from large geo tagged databases~\cite{arandjelovic2016netvlad,radenovic2018fine}. Performance remains limited by database scope, update frequency and record accuracy. Recent progress in Large Vision Language Models LVLMs opens new opportunities and enables direct location reasoning from image content~\cite{vivanco2023geoclip,han2025swarm}. These models excel in interpreting complex visual information and absorb rich geospatial knowledge during pre training on diverse multimodal data.

Despite these advances, individual LVLMs still face challenges in complex geo localization scenarios~\cite{han2025swarm}. A single model can struggle with fine grained features across culturally diverse regions and can favor training locations that appear more often in pre training corpora. The same model often produces inconsistent predictions in the presence of ambiguous visual cues or highly specialized geographic scenes. To mitigate these issues, researchers have proposed multi agent collaboration frameworks that combine several LVLMs with complementary strengths~\cite{wu2024autogen,hong2024metagpt,liang2024mad}. Current systems usually adopt independent exploration with simple result aggregation or sequential collaboration with a fixed interaction order~\cite{han2025swarm}. These strategies leverage model diversity for improved performance but they still lack robust mechanisms for resolution of conflicting predictions. Simple aggregation cannot assess prediction reliability or merge competing interpretations in a principled way. As a result, existing approaches miss the chance to use structured argumentation to expose reasoning flaws and refine predictions through iterative feedback~\cite{goeckner2024graph}.

\begin{figure}[t]
  \centering
  \includegraphics[width=\columnwidth]{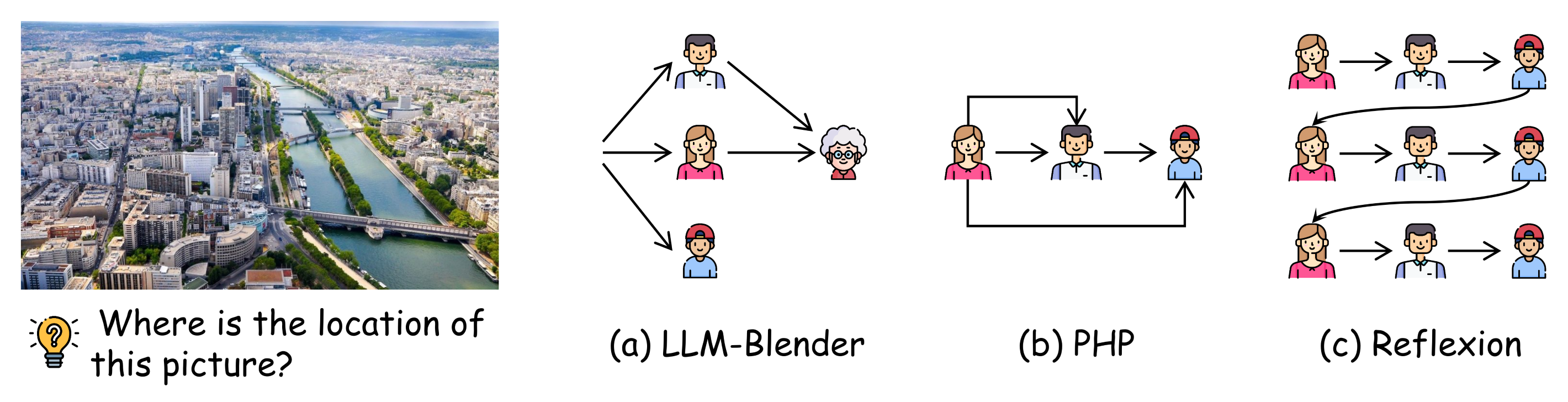}
  \caption{Comparison of multi-agent collaboration frameworks for visual geo-localization. (a) LLM-Blender: agents independently predict and aggregate through simple voting, lacking conflict resolution. (b) PHP: sequential collaboration with fixed order, prone to error propagation. (c) Reflexion: parallel self-reflection without inter-agent debate. These methods treat agent interactions uniformly and cannot effectively resolve conflicting predictions through structured argumentation.
}
  \label{figure1}
\end{figure}


A key technical challenge is to model the heterogeneous nature of agent interactions and to exploit this structure for conflict resolution. Different agents form distinct relationships based on prediction consistency and confidence. Agents with similar predictions benefit from supportive collaboration that consolidates shared evidence. Agents with conflicting predictions need competitive argumentation that challenges assumptions and reveals errors. Real world geo localization often contains such mixed signals since architectural elements can suggest one region and natural landscapes can indicate another location. Simple aggregation methods such as majority voting cannot measure prediction reliability or detect conflicts that reveal complementary evidence. Knowledge needs directional flow from high confidence agents toward uncertain agents. Human experts handle disagreements through explicit reasoning, presentation of evidence and careful examination of alternative hypotheses. They build consensus through structured debate that exposes flaws and synthesizes diverse perspectives. Many current graph neural networks assume homogeneous connectivity and apply the same message passing rule on all edges so they cannot express distinct modes of cooperation and competition among agents.

Another challenge arises from dynamic evolution of debate relationships during collaboration. Interaction roles such as agreement, conflict and knowledge transfer can change as discussion progresses. Two agents can move from disagreement to consensus when they gather additional evidence. Agents that appear aligned at first can later expose subtle differences that require deeper analysis. Standard graph neural networks usually update node representations but keep edge relationships fixed. As a result they cannot track how argument strength and reliability evolve across debate rounds. Interaction history carries useful signals for design of collaboration patterns. An effective framework needs explicit pairwise modeling of interactions and adaptive graph topology so both node states and edge relations benefit from ongoing debate.

To address these challenges, we propose GraphGeo, a multi agent debate framework for visual geo localization based on a heterogeneous graph neural network. The framework turns cognitive conflicts into useful signals through structured debate among LVLM agents. GraphGeo represents each agent as a node and encodes debate relations as typed edges in a heterogeneous graph. Agreement edges support consolidation of knowledge through mutual reinforcement. Conflict edges drive competitive argument through critical feedback. Transfer edges guide knowledge sharing from confident agents toward uncertain agents. We design a dual level debate mechanism. Node level refinement aggregates relation specific messages through tailored message passing and edge level modeling maintains learnable representations of pairwise argument states. These edge states evolve through recurrent updates and shape later node reasoning. We also introduce cross level topology refinement so graph structure and agent representations co evolve during debate. In summary our work makes three main contributions:
\begin{enumerate}
\item[$\bullet$] We formulate multi agent visual geo localization as structured debate on a heterogeneous graph with typed relations for agreement, conflict and knowledge transfer. This formulation provides a unified view of cooperation and competition in LVLM based multi agent systems.
\item[$\bullet$] We develop GraphGeo, an architecture with dual level debate and cross level topology refinement. Node representations and collaboration patterns evolve jointly and convert cognitive conflicts into accurate geo localization predictions.
\item[$\bullet$] Extensive experiments show that heterogeneous debate improves both accuracy and robustness over strong single model and multi agent baselines. This study highlights the importance of structured interaction design for future LVLM collaboration frameworks.
\end{enumerate}

\section{Preliminary}
\subsection{Problem Formulation}
Visual geo-localization aims to predict the geographic location $\mathcal{Y}$ of a given image $\mathcal{X}$ without relying on GPS metadata. We formulate this task within a multi-agent framework where $N$ large vision-language models serve as collaborative agents $\mathcal{V} = \{v_1, v_2, \ldots, v_N\}$. Each agent $v_i$ generates location predictions $\mathcal{Y}_{v_i}$ with confidence scores $c_{v_i} \in [0,1]$. The collaborative prediction process can be modeled as a graph $\mathcal{G} = (\mathcal{V}, \mathcal{E})$ where nodes represent agents and edges $\mathcal{E}$ encode their interaction patterns. The optimal prediction $\mathcal{Y}^*$ emerges from aggregating diverse agent perspectives weighted by their expertise and mutual influence within the graph structure.

\subsection{Graph Neural Networks for Agent Collaboration}
Graph neural networks provide a principled framework for learning representations through iterative message passing. Given the agent collaboration graph $\mathcal{G}$, each agent $v_i$ maintains a hidden state $\mathbf{h}_i^{(l)} \in \mathbb{R}^d$ at layer $l$. The standard message passing mechanism updates node representations by aggregating neighborhood information through
\begin{equation}
\mathbf{h}_i^{(l+1)} = \text{UPDATE}^{(l)}\left(\mathbf{h}_i^{(l)}, \mathbf{a}_i^{(l)}\right),
\end{equation}
\begin{equation}
\mathbf{a}_i^{(l)} = \text{AGGREGATE}^{(l)}\left(\{\mathbf{m}_{ji}^{(l)} \mid v_j \in \mathcal{N}(v_i)\}\right).
\end{equation}
where $\mathcal{N}(v_i)$ denotes the neighbors of agent $v_i$ and $\mathbf{m}_{ji}^{(l)}$ represents the message from agent $v_j$ to $v_i$ at layer $l$. However, this homogeneous treatment of agent interactions fails to capture the competitive and collaborative dynamics essential for effective geo-localization. Agents may produce conflicting predictions that require resolution through argumentation rather than simple averaging, and different images demand different collaboration patterns among agents.

\begin{figure*}[t]
  \centering
  \includegraphics[width=\textwidth]{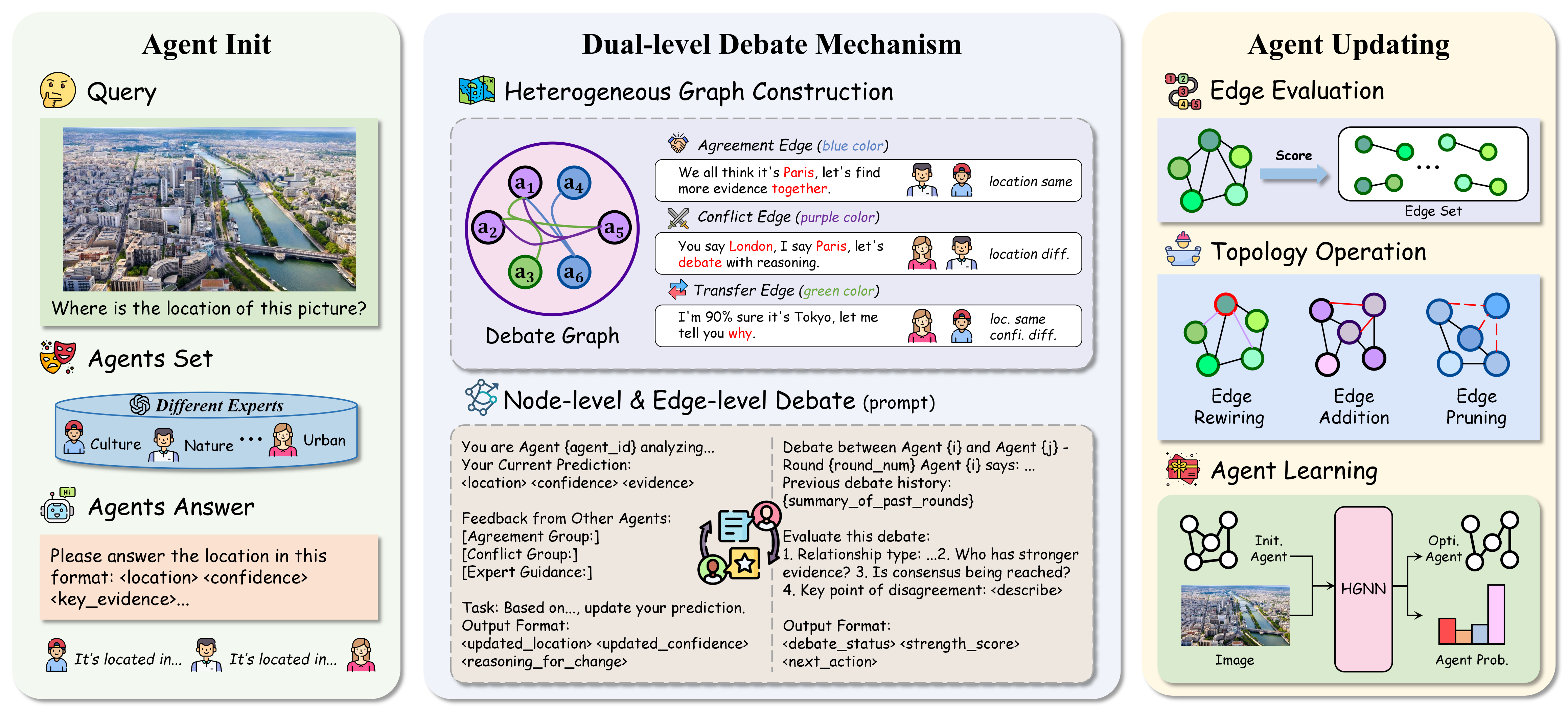}
  \caption{The framework overview of our proposed method. It contains a dual-level debate mechanism and cross-level topology refinement for visual geo-localization. The first part constructs a heterogeneous graph with three edge types to model diverse agent interactions. The second part employs node-level and edge-level debate to refine predictions through structured argumentation. The third part optimizes graph topology by evaluating edge utility and adjusting connections during multi-round debates.}
  \label{figure2}
\end{figure*}

\section{Methodology}
\subsection{Heterogeneous Graph Construction with Debate Edges}
We construct a heterogeneous agent collaboration graph that explicitly models diverse interaction types through typed edges. The graph $\mathcal{G} = (\mathcal{V}, \mathcal{E}, \mathcal{R})$ incorporates a set of relation types $\mathcal{R} = \{r_{\text{agree}}, r_{\text{conflict}}, r_{\text{transfer}}\}$ to distinguish between different debate mechanisms. For each image $\mathcal{X}$, agent nodes $v_i \in \mathcal{V}$ have features $\mathbf{x}_i^{(0)}$ that concatenate the VAE-encoded image representation with learnable agent embeddings $\mathbf{e}_i \in \mathbb{R}^{d_e}$. The edge set $\mathcal{E}$ is dynamically constructed based on initial agent predictions. We compute the geodesic distance $d_{\text{geo}}(\mathcal{Y}_{v_i}, \mathcal{Y}_{v_j})$ between predicted locations and assign relation types through
\begin{equation}
r_{ij} = \begin{cases}
r_{\text{agree}} & \text{if } d_{\text{geo}}(\mathcal{Y}_{v_i}, \mathcal{Y}_{v_j}) < \tau_{\text{agree}} \\
r_{\text{conflict}} & \text{if } d_{\text{geo}}(\mathcal{Y}_{v_i}, \mathcal{Y}_{v_j}) > \tau_{\text{conflict}} \\
r_{\text{transfer}} & \text{if } c_{v_i} - c_{v_j} > \tau_{\text{conf}}
\end{cases},
\end{equation}
where $\tau_{\text{agree}}$, $\tau_{\text{conflict}}$, and $\tau_{\text{conf}}$ are threshold hyperparameters. Agreement edges connect agents with consistent predictions to facilitate knowledge consolidation, conflict edges link agents with divergent views to enable competitive debate, and transfer edges direct information flow from high-confidence to low-confidence agents.

\subsection{Competitive Message Passing with Node-level Debate}
At each layer $l$, we replace standard message aggregation with a competitive debate mechanism. For each agent $v_i$, we partition its neighbors $\mathcal{N}(v_i)$ according to edge types as $\mathcal{N}_r(v_i) = \{v_j \mid (v_j, v_i, r) \in \mathcal{E}\}$ for $r \in \mathcal{R}$. The node-level debate generates relation-specific messages tailored to the interaction type. For agreement edges, agent $v_j$ generates supportive messages that reinforce shared understanding through
\begin{equation}
\mathbf{m}_{ji}^{\text{agree}} = \mathbf{W}_{\text{agree}} \cdot \sigma(\mathbf{h}_j^{(l)} \odot \mathbf{h}_i^{(l)}),
\end{equation}
where $\mathbf{W}_{\text{agree}} \in \mathbb{R}^{d \times d}$ is a learnable weight matrix and $\odot$ represents element-wise multiplication. For conflict edges, agent $v_j$ generates critical messages that challenge the current hypothesis through
\begin{equation}
\mathbf{m}_{ji}^{\text{conflict}} = \mathbf{W}_{\text{conflict}} \cdot \text{ReLU}(\mathbf{h}_j^{(l)} - \mathbf{h}_i^{(l)})
\end{equation}
Transfer edges employ an attention-based mechanism where high-confidence agent $v_j$ selectively transfers relevant knowledge through
\begin{equation}
\mathbf{m}_{ji}^{\text{transfer}} = \alpha_{ji} \mathbf{W}_{\text{transfer}} \mathbf{h}_j^{(l)},
\end{equation}
\begin{equation}
\alpha_{ji} = \frac{\exp(c_{v_j} \cdot \mathbf{h}_j^{(l)\top} \mathbf{W}_{\text{attn}} \mathbf{h}_i^{(l)})}{\sum_{v_k \in \mathcal{N}_{\text{transfer}}(v_i)} \exp(c_{v_k} \cdot \mathbf{h}_k^{(l)\top} \mathbf{W}_{\text{attn}} \mathbf{h}_i^{(l)})}
\end{equation}
where $\alpha_{ji}$ weights the transfer by both confidence $c_{v_j}$ and semantic relevance. The agent $v_i$ then aggregates these heterogeneous messages through relation-aware pooling
\begin{equation}
\mathbf{m}_i^{(l)} = \bigoplus_{r \in \mathcal{R}} \text{POOL}\left(\{\mathbf{m}_{ji}^r \mid v_j \in \mathcal{N}_r(v_i)\}\right),
\end{equation}
where $\bigoplus$ denotes a learnable combination operator that balances different debate mechanisms.

\subsection{Edge-level Debate through Cross-agent Interaction}
Beyond node-level refinement, we introduce edge-level debate that explicitly models pairwise interactions to capture fine-grained argumentation dynamics. For each edge $(v_i, v_j, r) \in \mathcal{E}$, we construct a debate representation $\mathbf{d}_{ij}^{(l)}$ that encodes the interaction state between agents. This edge representation evolves through
\begin{equation}
\mathbf{d}_{ij}^{(l)} = \mathbf{W}_{\text{edge}}^r \cdot [\mathbf{h}_i^{(l)} \oplus \mathbf{h}_j^{(l)} \oplus \Delta\mathcal{Y}_{ij}],
\end{equation}
\begin{equation}
\mathbf{d}_{ij}^{(l+1)} = \text{GRU}\left(\mathbf{d}_{ij}^{(l)}, \mathbf{h}_i^{(l+1)} + \mathbf{h}_j^{(l+1)}\right),
\end{equation}
where $\mathbf{W}_{\text{edge}}^r \in \mathbb{R}^{d_e \times 3d}$ is a relation-specific projection matrix and GRU maintains debate history while incorporating updated node states. The edge representations subsequently influence node updates through edge-to-node message passing
\begin{equation}
\mathbf{m}_i^{\text{edge},(l)} = \sum_{v_j \in \mathcal{N}(v_i)} \beta_{ij} \mathbf{d}_{ij}^{(l)}, 
\end{equation}
\begin{equation}
\beta_{ij} = \text{softmax}_j(\mathbf{d}_{ij}^{(l)\top} \mathbf{W}_{\text{attend}} \mathbf{h}_i^{(l)})
\end{equation}
The final node update integrates both node-level and edge-level messages through a gated fusion mechanism
\begin{equation}
\mathbf{h}_i^{(l+1)} = \text{GRU}\left(\mathbf{h}_i^{(l)}, \gamma_i \mathbf{m}_i^{(l)} + (1-\gamma_i) \mathbf{m}_i^{\text{edge},(l)}\right),
\end{equation}
where the gate $\gamma_i = \sigma(\mathbf{w}_{\gamma}^\top [\mathbf{h}_i^{(l)} \oplus \mathbf{m}_i^{(l)} \oplus \mathbf{m}_i^{\text{edge},(l)}])$ adaptively balances the two information sources.

\subsection{Cross-level Topology Refinement}
To enable dynamic adaptation of the debate structure, we introduce a cross-level topology refinement mechanism. After $L$ layers of message passing, we obtain final agent representations $\mathbf{H} = \{\mathbf{h}_i^{(L)} \mid v_i \in \mathcal{V}\}$ that encode refined predictions. These representations feedback to the graph topology through a learned edge scoring function
\begin{equation}
s_{ij} = \text{MLP}_{\text{score}}\left([\mathbf{h}_i^{(L)} \oplus \mathbf{h}_j^{(L)} \oplus \mathbf{d}_{ij}^{(L)} \oplus \Delta\mathbf{h}_{ij}]\right),
\end{equation}
where $\Delta\mathbf{h}_{ij} = \mathbf{h}_i^{(L)} - \mathbf{h}_i^{(0)}$ captures the representational change induced by debate. Edges with low utility scores are pruned while new edges are added between agents whose representations become similar after debate. The topology learning objective combines task-specific loss with graph regularization
\begin{equation}
\mathcal{L}_{\text{total}} = \mathcal{L}_{\text{geo}} + \lambda_{\text{sparse}} \sum_{(i,j) \in \mathcal{E}} s_{ij} + \lambda_{\text{diverse}} \mathcal{H}(\mathcal{R}),
\end{equation}
where $\mathcal{L}_{\text{geo}}$ measures the geodesic distance between predicted and ground-truth locations, the sparsity term penalizes excessive edge connectivity, and $\mathcal{H}(\mathcal{R})$ represents the entropy over relation type distribution. This cross-level interaction ensures that the topology evolves to support effective collaboration patterns while node representations adapt to leverage the refined debate structure.

\begin{table}[t]
\caption{Results of different single LVLM baselines and comparison with baselines. Bold indicates the statistically significant improvements over the best baseline.}
\label{table1}
\centering
\renewcommand\tabcolsep{2.5pt}
\begin{tabular}{ccccc}
\hline
\specialrule{0em}{1pt}{1pt}
\multirow{2}{*}{\bfseries{Model}} & \multirow{2}{*}{\bfseries{IM2GPS3K}} & \multirow{2}{*}{\bfseries{YFCC4K}} & \bfseries{GeoGlobe} & \bfseries{GeoGlobe} \\ & & & \bfseries{(Natural)} & \bfseries{(ManMade)}
\\ \specialrule{0em}{1pt}{1pt}
\hline
\multicolumn{5}{c}{\textit{Open-source LVLMs}} \\
\hline
Infi-MM & 14.7000 & 6.5256 & 19.2020 & 21.4145 \\
Qwen-VL & 32.4667 & 14.3959 & 42.3940 & 37.4556 \\
Qwen2-VL-7B & 36.2333 & 15.8724 & 45.6783 & 52.3142 \\
InternVL2-8B & 33.5667 & 14.9213 & 40.1256 & 48.7623 \\
vip-llava-13b & 15.7667 & 6.9885 & 20.6983 & 15.4089 \\
vip-llava-7b & 38.7000 & 17.8792 & 31.9202 & 56.4994 \\
llava-1.5-7b & 30.2000 & 13.8889 & 27.1820 & 47.2145 \\
llava-1.6-7b-mistral & 3.1000 & 1.6314 & 0.7481 & 2.1731 \\
llava-1.6-7b-vicuna & 9.4667 & 4.4092 & 6.9825 & 15.8831 \\
llava-1.6-13b & 17.8333 & 8.3333 & 12.2195 & 28.2497 \\
llava-1.6-34b & 44.2667 & 20.8113 & 52.8678 & 77.2027 \\
CogVLM & 6.7333 & 3.0644 & 7.7307 & 10.3516 \\
CogVLM2 & 28.4333 & 12.5678 & 35.4521 & 42.3187 \\
\hline
\multicolumn{5}{c}{\textit{Proprietary LVLMs}} \\
\hline
Claude-3-Opus & 23.8333 & 12.3457 & 33.1671 & 40.6954 \\
Claude-3.5-Sonnet & 41.2667 & 18.6735 & 55.3214 & 76.8456 \\
GPT-4V & 43.5333 & 19.2147 & 58.4672 & 80.1234 \\
GPT-4o & 46.8333 & 20.4569 & 64.7823 & 85.2617 \\
GPT-4o-mini & 45.0000 & 18.3422 & 62.0948 & 84.5911 \\
Gemini-1.5-Flash & 39.4667 & 17.2341 & 54.8912 & 74.5623 \\
Gemini-1.5-Pro & 47.3667 & 19.9956 & 62.3441 & 82.8131 \\
\hline
\multicolumn{5}{c}{\textit{Visual Geo-localization Methods}} \\
\hline
NetVLAD & 16.6303 & 7.4876 & 26.5134 & 28.9955 \\
GeM & 14.4907 & 6.5243 & 23.1022 & 25.4175 \\
CosPlace & 17.6686 & 7.9551 & 28.1688 & 30.2782 \\
PIGEON & 29.5632 & 12.8914 & 35.7891 & 41.2365 \\
StreetCLIP & 33.2145 & 14.6782 & 39.5621 & 44.8912 \\
GeoCLIP & 34.4728 & 15.1719 & 38.2519 & 45.9174 \\
smileGeo & 47.7667 & 21.5168 & 76.0599 & 85.4603 \\
\hline
\textbf{Ours} & \textbf{49.1456} & \textbf{22.1623} & \textbf{77.8217} & \textbf{87.9341} \\
\hline
\end{tabular}
\end{table}

\section{Experiments}
\subsection{Datasets}
We evaluate our approach on four geo-localization benchmarks. The \textbf{IM2GPS3K} dataset \cite{vo2017revisiting} contains 3,000 images sampled from the original IM2GPS collection, covering multiple continents with urban, natural, and landmark scenes. The \textbf{YFCC4K} dataset \cite{vo2017revisiting} consists of 4,536 geo-tagged images from Yahoo Flickr Creative Commons 100 Million collection with broad visual and geographic diversity. We also utilize the \textbf{GeoGlobe} dataset \cite{han2025swarm}, divided into \textbf{GeoGlobe (Natural)} for natural attractions and \textbf{GeoGlobe (ManMade)} for man-made landmarks. This dataset filters out ambiguous images and provides clearer localization targets across different regions.

\begin{figure*}[t]
  \centering
  \includegraphics[width=\textwidth]{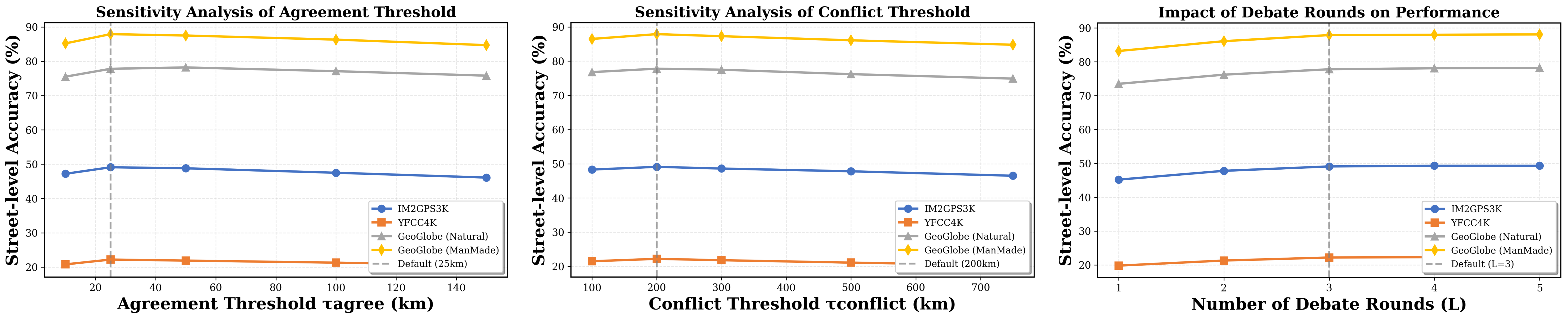}
\caption{Hyperparameter sensitivity analysis. (a) Agreement threshold $\tau_{agree}$ shows optimal values vary across datasets due to different geographic distributions, with default at 25 km. (b) Conflict threshold $\tau_{conflict}$ balances critical debate and noise, where lower values create more conflict edges. (c) Debate rounds show performance gains diminish after 3 rounds, making $L=3$ optimal for accuracy-cost trade-off.}
  \label{figure3}
\end{figure*}
\begin{figure*}[t]
  \centering
  \includegraphics[width=\textwidth]{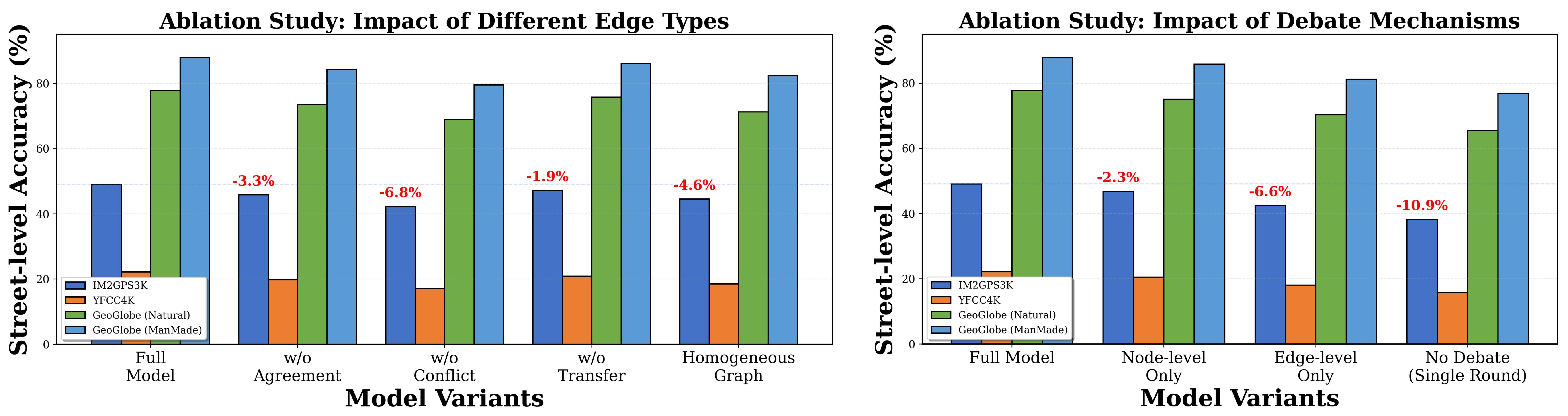}
\caption{Ablation Study on Edge Types and Debate Mechanism
Caption: Ablation studies on heterogeneous edge types and dual-level debate mechanism. Conflict edges are most critical, with 6.8\% performance drop when removed, highlighting the essential role of competitive debate in error correction. Node-level aggregation proves more critical than edge-level modeling, while removing the entire debate mechanism causes 10.9\% performance drop on IM2GPS3K, validating the core value of iterative multi-agent argumentation.
}
  \label{figure4}
\end{figure*}

\subsection{Baselines}
We compare our proposed method with three categories of baselines. First, we evaluate \textbf{single LVLM agents} including both open-source models (Qwen-VL \cite{bai2023qwen}, Qwen2-VL \cite{wang2024qwen2vl}, InternVL2 \cite{chen2024expanding}, LLaVA series \cite{liu2023llava}, CogVLM \cite{wang2024cogvlm}) and proprietary models (Claude-3.5-Sonnet \cite{anthropic2024claude}, GPT-4 series \cite{openai2024gpt4}, Gemini series \cite{team2024gemini}) to establish individual model performance. Second, we include \textbf{traditional visual geo-localization methods} such as NetVLAD \cite{arandjelovic2016netvlad}, which pioneered trainable VLAD pooling, GeM \cite{radenovic2018fine} using generalized mean pooling, CosPlace \cite{berton2022cosplace} reformulating the task as classification for scalability, PIGEON \cite{haas2024pigeon}, StreetCLIP \cite{haas2023streetclip}, and GeoCLIP \cite{vivanco2023geoclip} integrating semantic reasoning into features. Third, we compare with \textbf{smileGeo}\cite{han2025swarm}, a swarm intelligence framework aggregating multiple LVLM agents through collaborative reasoning, representing the current state-of-the-art in multi-agent geo-localization.

\subsection{Metrics}
Following common practice in visual geo localization, we adopt street level accuracy as the main evaluation metric. It measures the percentage of images localized within a distance threshold from the ground truth. We report accuracy at several scales from fine street level at $1$ km to coarse continent level at $2500$ km. For each dataset, we compute the proportion of test images whose predicted locations lie inside the given radius around the true coordinates. This metric is easy to interpret and supports fair comparison across methods and datasets.

\subsection{Implementation Details}
We implement all experiments on NVIDIA A100 GPUs. Our multi-agent system incorporates a diverse set of LVLM agents selected from both open-source and proprietary models to ensure complementary reasoning capabilities. For graph construction, we initialize the heterogeneous graph with typed edges based on prediction similarity, setting distance thresholds $\tau_{\text{agree}} = 25$km, $\tau_{\text{conflict}} = 200$km, and $\tau_{\text{conf}} = 0.3$ for edge type assignment. The dual-level debate mechanism operates through $L = 3$ layers of message passing with hidden dimension $d = 512$. We employ the Adam optimizer with learning rate $1\times10^{-4}$ and train for 50 epochs with batch size 32. The cross-level topology refinement uses hyperparameters $\lambda_{\text{sparse}} = 0.01$ and $\lambda_{\text{diverse}} = 0.05$ to balance prediction accuracy with graph structure quality. During inference, we aggregate final agent predictions through confidence-weighted voting, where each agent's contribution is determined by its refined confidence score after multi-round debate.

\subsection{Main Results and Model Analysis}
Table \ref{table1} presents the results of our method and all baselines including open source LVLMs, proprietary models, traditional geo localization methods, and the multi agent system smileGeo. Our method attains the best performance on all four benchmarks with street level accuracy of $49.1\%$ on IM2GPS3K and $87.9\%$ on GeoGlobe ManMade and it surpasses the strongest baseline smileGeo on both IM2GPS3K and GeoGlobe by clear margins. Proprietary LVLMs show strong single model ability and open source LVLMs display larger variance in accuracy and traditional geo localization methods remain far behind vision language models, so semantic reasoning becomes a key factor for this task. The multi agent baseline smileGeo already improves over every single model through collaborative reasoning yet Our method brings further gains. The core idea behind Our method is that interactions among agents are heterogeneous and need different handling through typed edges. Agreement edges support collaboration and knowledge consolidation, conflict edges drive competitive argument and error discovery, and transfer edges guide knowledge flow from confident agents to uncertain agents. The dual level debate mechanism combines node level message aggregation with edge level interaction modeling so agents can absorb diverse views and keep explicit debate states that evolve across rounds.

\subsection{Hyper-parameter Analysis}
Fig.~\ref{figure3} examines the sensitivity of key hyperparameters across our four benchmarks. The agreement threshold shows optimal values vary by dataset, with 25km working best overall as it balances supportive collaboration without over-connecting dissimilar predictions. Too small values create overly sparse agreement graphs that limit beneficial consolidation. The conflict threshold at 200km effectively separates genuine disagreements from noise. Too large values blur the distinction between conflicting and merely distant predictions. Debate rounds show performance gains through three iterations, after which improvements plateau as agents reach consensus. The dynamic nature of our graph topology allows the framework to adapt edge connections based on debate effectiveness, with cross-level refinement enabling bidirectional influence between structure and representations. This adaptive mechanism ensures that productive debate relationships strengthen over time while unproductive connections are pruned.

\subsection{Ablation Study}
Fig.~\ref{figure4} systematically evaluates each component's contribution to our framework. Removing conflict edges causes the largest performance drop of 6.8\% on IM2GPS3K, highlighting that competitive argumentation is essential for error correction. Agreement edges contribute 3.3\% and transfer edges add 1.9\%, while the homogeneous graph variant loses 4.6\% by treating all interactions uniformly. Eliminating all iterative refinement causes a dramatic 10.9\% performance drop, validating that multi-round argumentation forms the core value of our approach. The results reveal that different agent relationships serve fundamentally different purposes in collaborative geo-localization. Agreement requires consolidation of shared understanding, conflict requires structured channels for challenging assumptions, and confidence gaps require directional knowledge flow. Our heterogeneous graph design captures these distinct interaction semantics, enabling agents to leverage both cooperation and competition.

\section{Conclusion}
This paper introduces GraphGeo, a multi-agent debate framework for visual geo-localization using heterogeneous graph neural networks. By modeling diverse agent interactions as typed edges and implementing dual-level debate with cross-level topology refinement, our framework turns cognitive conflicts into enhanced localization accuracy. Results demonstrate that heterogeneous interaction modeling and structured debate are crucial for multi-agent consensus under uncertainty.


\bibliographystyle{IEEEtran}
\bibliography{icme2026references}

@inproceedings{hays2008im2gps,
  title={im2gps: estimating geographic information from a single image},
  author={Hays, James and Efros, Alexei A},
  booktitle={Proceedings of the IEEE Conference on Computer Vision and Pattern Recognition},
  year={2008}
}

@inproceedings{baatz2012large,
  title={Large scale visual geo-localization of images in mountainous terrain},
  author={Baatz, Georges and Saurer, Olivier and K{\"o}ser, Kevin and Pollefeys, Marc},
  booktitle={Computer Vision--ECCV 2012: 12th European Conference on Computer Vision, Florence, Italy, October 7-13, 2012, Proceedings, Part I 12},
  pages={517--530},
  year={2012},
  organization={Springer}
}

@inproceedings{arandjelovic2016netvlad,
  title={{NetVLAD}: {CNN} architecture for weakly supervised place recognition},
  author={Arandjelovic, Relja and Gronat, Petr and Torii, Akihiko and Pajdla, Tomas and Sivic, Josef},
  booktitle={Proceedings of the IEEE Conference on Computer Vision and Pattern Recognition},
  pages={5297--5307},
  year={2016}
}

@inproceedings{vo2017revisiting,
  title={Revisiting IM2GPS in the deep learning era},
  author={Vo, Nam and Jacobs, Nathan and Hays, James},
  booktitle={Proceedings of the IEEE International Conference on Computer Vision},
  pages={2621--2630},
  year={2017}
}

@article{radenovic2018fine,
  author={Radenović, Filip and Tolias, Giorgos and Chum, Ondřej},
  journal={IEEE Transactions on Pattern Analysis and Machine Intelligence}, 
  title={Fine-Tuning CNN Image Retrieval with No Human Annotation}, 
  year={2019},
  volume={41},
  number={7},
  pages={1655-1668}
}

@article{vivanco2023geoclip,
  title={Geoclip: Clip-inspired alignment between locations and images for effective worldwide geo-localization},
  author={Vivanco Cepeda, Vicente and Nayak, Gaurav Kumar and Shah, Mubarak},
  journal={Advances in Neural Information Processing Systems},
  volume={36},
  pages={8690--8701},
  year={2023}
}

@inproceedings{han2025swarm,
  title={Swarm intelligence in geo-localization: A multi-agent large vision-language model collaborative framework},
  author={Han, Xiao and Zhu, Chen and Zhu, Hengshu and Zhao, Xiangyu},
  booktitle={Proceedings of the 31st ACM SIGKDD Conference on Knowledge Discovery and Data Mining V. 2},
  pages={814--825},
  year={2025}
}

@inproceedings{bai2023qwen,
  title={Qwen-VL: A Versatile Vision-Language Model for Understanding, Localization, Text Reading, and Beyond},
  author={Bai, Jinze and Bai, Shuai and Yang, Shusheng and Wang, Shijie and Tan, Sinan and Wang, Peng and Lin, Junyang and Zhou, Chang and Zhou, Jingren},
  booktitle={arXiv preprint arXiv:2308.12966},
  year={2023}
}

@article{liu2023llava,
  title={Visual instruction tuning},
  author={Liu, Haotian and Li, Chunyuan and Wu, Qingyang and Lee, Yong Jae},
  journal={Advances in Neural Information Processing Systems},
  volume={36},
  pages={34892--34916},
  year={2023}
}

@article{wang2024cogvlm,
  title={CogVLM: Visual Expert for Pretrained Language Models},
  author={Wang, Weihan and Lv, Qingsong and Yu, Wenmeng and Hong, Wenyi and Qi, Ji and Wang, Yan and Ji, Junhui and Yang, Zhuoyi and Zhao, Lei and Song, Xixuan and others},
  journal={Advances in Neural Information Processing Systems},
  volume={37},
  pages={121475--121499},
  year={2024}
}

@misc{anthropic2024claude,
  title={The Claude 3 Model Family: Opus, Sonnet, Haiku},
  author={Anthropic},
  howpublished={\url{https://www.anthropic.com/claude}},
  year={2024}
}

@article{team2024gemini,
  title={Gemini 1.5: Unlocking multimodal understanding across millions of tokens of context},
  author={Team, Gemini and Georgiev, Petko and Lei, Ving Ian and Burnell, Ryan and Bai, Libin and Gulati, Anmol and Tanzer, Garrett and Vincent, Damien and Pan, Zhufeng and Wang, Shibo and others},
  journal={arXiv preprint arXiv:2403.05530},
  year={2024}
}

@article{wang2024qwen2vl,
  title={Qwen2-VL: Enhancing Vision-Language Model's Perception of the World at Any Resolution},
  author={Wang, Peng and Bai, Shuai and Tan, Sinan and Wang, Shijie and Fan, Zhihao and Bai, Jinze and Chen, Keqin and Liu, Xuejing and Wang, Jialin and Ge, Wenbin and others},
  journal={arXiv preprint arXiv:2409.12191},
  year={2024}
}

@article{chen2024expanding,
  title={Expanding performance boundaries of open-source multimodal models with model, data, and test-time scaling},
  author={Chen, Zhe and Wang, Weiyun and Cao, Yue and Liu, Yangzhou and Gao, Zhangwei and Cui, Erfei and Zhu, Jinguo and Ye, Shenglong and Tian, Hao and Liu, Zhaoyang and others},
  journal={arXiv preprint arXiv:2412.05271},
  year={2024}
}

@inproceedings{haas2024pigeon,
  title={Pigeon: Predicting image geolocations},
  author={Haas, Lukas and Skreta, Michal and Alberti, Silas and Finn, Chelsea},
  booktitle={Proceedings of the IEEE/CVF Conference on Computer Vision and Pattern Recognition},
  pages={12893--12902},
  year={2024}
}

@article{haas2023streetclip,
title={Learning Generalized Zero-Shot Learners for Open-Domain Image Geolocalization},
author={Haas, Lukas and Alberti, Silas and Skreta, Michal},
journal={arXiv preprint arXiv:2302.00275},
year={2023}
}

@inproceedings{wu2024autogen,
  title={Autogen: Enabling next-gen LLM applications via multi-agent conversations},
  author={Wu, Qingyun and Bansal, Gagan and Zhang, Jieyu and Wu, Yiran and Li, Beibin and Zhu, Erkang and Jiang, Li and Zhang, Xiaoyun and Zhang, Shaokun and Liu, Jiale and others},
  booktitle={First conference on language modeling},
  year={2024}
}

@inproceedings{hong2024metagpt,
  title={Meta{GPT}: Meta programming for a multi-agent collaborative framework},
  author={Hong, Sirui and Zhuge, Mingchen and Chen, Jonathan and Zheng, Xiawu and Cheng, Yuheng and Wang, Jinlin and Zhang, Ceyao and Wang, Zili and Yau, Steven Ka Shing and Lin, Zijuan and others},
  booktitle={The Twelfth International Conference on Learning Representations},
  year={2024}
}

@inproceedings{liang2024mad,
  title={Encouraging divergent thinking in large language models through multi-agent debate},
  author={Liang, Tian and He, Zhiwei and Jiao, Wenxiang and Wang, Xing and Wang, Yan and Wang, Rui and Yang, Yujiu and Shi, Shuming and Tu, Zhaopeng},
  booktitle={Proceedings of the 2024 Conference on Empirical Methods in Natural Language Processing},
  pages={17889--17904},
  year={2024}
}

@inproceedings{berton2022cosplace,
  title={Rethinking Visual Geo-localization for Large-Scale Applications},
  author={Berton, Gabriele and Masone, Carlo and Caputo, Barbara},
  booktitle={Proceedings of the IEEE/CVF Conference on Computer Vision and Pattern Recognition},
  pages={4878--4888},
  year={2022}
}

@inproceedings{goeckner2024graph,
  title={Graph neural network-based multi-agent reinforcement learning for resilient distributed coordination of multi-robot systems},
  author={Goeckner, Anthony and Sui, Yueyuan and Martinet, Nicolas and Li, Xinliang and Zhu, Qi},
  booktitle={2024 IEEE/RSJ International Conference on Intelligent Robots and Systems (IROS)},
  pages={8586--8593},
  year={2024},
  organization={IEEE}
}

@article{openai2024gpt4,
  title={Gpt-4 technical report},
  author={Achiam, Josh and Adler, Steven and Agarwal, Sandhini and Ahmad, Lama and Akkaya, Ilge and Aleman, Florencia Leoni and Almeida, Diogo and Altenschmidt, Janko and Altman, Sam and Anadkat, Shyamal and others},
  journal={arXiv preprint arXiv:2303.08774},
  year={2023}
}

\end{document}